\documentclass{article}

\usepackage[numbers,sort]{natbib}
\usepackage[hyphens]{url}

\usepackage[multiple]{footmisc}
\usepackage{multirow}
\usepackage{booktabs}
\usepackage[np,newcolumntypes]{numprint}
\npthousandsep{,}
\npproductsign{\times}
\npdecimalsign{.}

\usepackage[latin1]{inputenc}   
\usepackage[T1]{fontenc}   

\usepackage{tikz}

\usepackage{enumitem}
\usepackage{natbib}
\newcommand{\citepos}[1]{\citeauthor{#1}'s \citep{#1}}
\usepackage{xspace}
\usepackage{soul}
\newcommand{\fsl}{\textsl}
\usepackage{xcolor}

\newcommand{\corpus}{\textsc{Paco}\xspace}

\newcommand{\mps}[1]{}

\newcounter{textboxno}
\usepackage[many]{tcolorbox}
\newtcolorbox{mybox}[2][]{%
boxsep=3pt,left=2pt,right=2pt,bottom=5pt,
width=\columnwidth,
boxrule=1pt,
attach boxed title to top center = {yshift=-\tcboxedtitleheight/2},
colbacktitle=white,coltitle=black,
boxed title style={size=normal,colframe=white,boxrule=0pt}, 
interior style={white},
title={\refstepcounter{textboxno}\label{#1}
Example \arabic{textboxno}: {#2}
\def\@currentlabel{\p@textboxno\thetextboxno}},
enhanced,
float,
}

\usepackage{soul}

\usepackage[normalem]{ulem}

\begin{document}

\title{\corpus : Provocation Involving Action, Culture, and Oppression}

\author{Vaibhav Garg\textsuperscript{\textsection},
Ganning Xu\textsuperscript{\textsection} 
and Munindar P. Singh 
\\
vgarg3@ncsu.edu,
ganningxu@gmail.com,
mpsingh@ncsu.edu
}



\maketitle
\pagestyle{plain}
\begingroup\renewcommand\thefootnote{\textsection}
\footnotetext{Garg and Xu made equal contributions to this work.}
\endgroup
\begin{abstract}

\emph{Warning: This paper may include examples that can be triggering to some readers, especially to the people of a religious group.}

In India, people identify with a particular group based on certain attributes such as religion. The same religious groups are often provoked against each other. Previous studies show the role of provocation in increasing tensions between India's two prominent religious groups: Hindus and Muslims. With the advent of the Internet, such provocation also surfaced on social media platforms such as WhatsApp. 

By leveraging an existing dataset of Indian WhatsApp posts, we identified three categories of provoking sentences against Indian Muslims. Further, we labeled \np{7000} sentences for three provocation categories and called this dataset \corpus. We leveraged \corpus to train a model that can identify provoking sentences from a WhatsApp post. Our best model is fine-tuned RoBERTa and achieved \np{0.851} average AUC score over five-fold cross-validation. Automatically identifying provoking sentences could stop provoking text from reaching out to the masses, and can prevent possible discrimination or violence against the target religious group.

Further, we studied the provocative speech through a pragmatic lens, by identifying the dialog acts and impoliteness super-strategies used against the religious group.

\end{abstract}

\section{Introduction}

In India, most people identify themselves with a religion, leading to two main groups: Hindus and Muslims, who form 79\% and 13\% of Indians, respectively \cite{Fearspeech2021}. However, the same identifying attribute, religion, is often misused to provoke these groups against each other \cite{Religious-provocation}. 

In psychology, provocation is considered antecedent to aggression and violence \citep{Aggressionprovocation2002,Violenceprovocation2013}. Hence, in our setting, we define \emph{provoking sentences} as those that either make the readers angry (at a religious group) or urge them to take any action (against that religious group). \citet{Fearspeech2021} curate a dataset of WhatsApp posts that instill fear of Indian Muslims in the mind of readers. Our investigation found that such fearful posts also contain provoking sentences against the same religious group. Example~\ref{box:fear-provocation} shows one such WhatsApp post. The sentence colored blue may instill fear among the readers, especially among citizens of the mentioned states due to the possible Islamic acquisition. Whereas, its subsequent sentences (colored red) typecast Muslims and can make the readers angry at them. The readers of such posts can be Muslims themselves but the readers who are provoked belong to other religious groups.

\begin{mybox}[box:fear-provocation]{Provocation accompanied with fear}
``Leave chatting and read this post or else all your life will be left in chatting. In 1378, a part was separated from India, became an Islamic nation - named Iran \ldots and \textcolor{blue}{now Uttar Pradesh, Assam and Kerala are on the verge of becoming an Islamic state} \ldots \textcolor{red}{People who do love jihad is a Muslim. Those who think of ruining the country - Every single one of them is a Muslim !!!! Everyone who does not share this message forward should be a Muslim.} \ldots''
\end{mybox}

Our manual investigation of WhatsApp posts reveals the following three types of provoking sentences:
\begin{description}
\item[Provocation involving religious culture:] Sentences that can make readers annoyed at a religion's scriptures, religious practices, or religious leaders fall in this category. Moreover, the sentences stereotyping all people of that religion are also considered relevant.

\item[Provocation involving religious oppression:] Sentences highlighting past misdeeds (either real incidents or fake) of a religious group, such as its violence, domination, or superiority over others.

\item[Provocation involving action:] Sentences that urge its readers to act against a religious group. Such actions include violence, discriminating against people of that religion, or boycotting that religious group or their monuments. 
\end{description}

\begin{mybox}[box:provocation-types]{Categories of provoking sentences}
\textbf{Religious culture}\\
``\ldots Hindu-Muslim unity is impossible because the Muslim Quran does not tolerate Hinduism as a friend.''\bigskip

\textbf{Religious oppression}\\
``\ldots Only Hindu temples are destroyed and business and other activities are being increased in their place and Hindus are also being attacked!''\bigskip

\textbf{Action}\\
``Hindu society should stop worshiping these tombstones, tombs, pirs.''
\end{mybox}

Example~\ref{box:provocation-types} shows a sentence for each type of provocation. The first sentence can make readers angry at the religious book, Quran (Islamic scripture). We consider all the sentences that target scriptures, leaders, or people of religion as provoking involving religious culture. The second sentence provokes readers by mentioning the past misdeeds of Muslims destroying Hindu temples, hence it falls under the religious oppression category. Whereas, the third sentence asks Hindu readers to not worship tombstones of other religions (Islam according to the context of the post), making it provocation involving action. 

\textbf{Difference from hate speech:} Hate speech generally leverages derogatory keywords to take direct digs at the target \cite{Hatekeywords2018}, which is not necessary for provocative speech. The three sentences in Example~\ref{box:provocation-types} show how readers can be provoked without using abusive and derogatory words. Due to this indirect nature of provocative speech, the identification of provoking sentences is even more difficult. 

Provoking sentences can be disturbing for the target religious group (Muslims in our case) and can emotionally harm them, especially on platforms such as WhatsApp where there is no content moderation due to end-to-end encryption \cite{Fearspeech2021}. On the other hand, non-target groups can end up sharing posts containing provoking sentences. This can even lead to violence against the target religious group \cite{Violenceprovocation2013}.  Therefore, we propose the following research questions.

\subsection{Research Questions}

\begin{description}
\item[RQ\textsubscript{identify}:] How can we automatically identify three types of provoking sentences from a WhatsApp post?
\end{description} 

Since a post can contain multiple types of provoking sentences, we target sentence-level classification in RQ\textsubscript{identify}. Moreover, sentence-level classification can help in pinpointing the sentences (from the whole post) that are responsible for provoking readers.

\begin{description}
\item[RQ\textsubscript{pragmatics}:] What dialogue acts and impoliteness super-strategies are used in provocative speech?
\end{description} 

To understand provoking sentences from an illocutionary perspective, identifying dialogue acts is important \citep{Speechacts1962,Speechacts1969,DAmodeling2000}. A dialog act is defined as the function of the speaker's utterance and hence reflects their intention behind the dialog \citep{Speechacts1962,Speechacts1969,DAmodeling2000}. Recent social media research stress on identifying dialog acts specific to the scenario \citep{Metooma2020,Twitterdialog2015}. Therefore, through our qualitative analysis, we find the dialog acts used in the provocation (against Muslims) scenario.

Provoking sentences involving religious culture and oppression can attack Muslims, their scriptures, their leaders, and so on. As a result, such sentences are impolite to Muslim readers. We identify what impoliteness super-strategies are commonly used in these sentences. Identifying dialogue acts and impoliteness super-strategies provides a pragmatic understanding of the provocative text against Indian Muslims.

\subsection{Contributions and Novelty}
We make the following contributions:

\begin{itemize}
    \item \corpus, a dataset of \np{7000} sentences annotated for three provocation categories: religious culture, religious oppression, and action. 
    
    \item On \corpus, We train Natural Language Processing (NLP) model to identify provoking sentences from a post. 

    \item We identify the characteristics of provocative language used in the religious context. To do so, we uncover dialog acts and impoliteness super-strategies used by the writer of such text.
    
\end{itemize}

Prior studies focus on either identifying hate speech or fear speech \cite{Fearspeech2021, Multilingualhatemodel2021, Multilingualabusivemodel2022, Hatelexicon2018} but not provocative speech. To the best of our knowledge, we are the first ones to computationally identify provocative speech in the religious context. We not only identify provoking sentences (against Muslims) but also study the pragmatics of such provocative language through dialogue acts and impoliteness super-strategies. 

\subsection{Key Findings}

Prior studies \cite{Fearspeech2021, Multilingualhatemodel2021, Multilingualabusivemodel2022, Hatelexicon2018} focused on identifying hate or fear speech but not provocative speech against a religious group.
On the other hand, we trained a transformer-based model that  achieved an average of 0.851 AUC-ROC score (Area Under the Receiver Operating Characteristic curve) for five-fold cross validation of \corpus.

Our qualitative analysis revealed six types of dialog acts that are used in provocative speech against Muslims. They are as follows: \emph{ 1) Accusation,  2) Defaming Muslims, 3) Criticizing Islam, 4) Comparison, 5) Commanding, and 6) Motivating}.
Moreover, we leveraged \citepos{Impolitenessmodel2003} 
 impoliteness model and found that negative impoliteness and bald-on-record are the most prominent super-strategies used across culture and oppression categories.

\subsection{Paper Organization}
Section~\ref{sec:relatedwork} lists the related work on hate and fear speech, and discusses how provocative speech is different from both of them. 
Section~\ref{sec:rq} describes the steps that we take to address RQ\textsubscript{identify}, including curation of \corpus and training multiple models. Section~\ref{sec:rq2} describes the qualitative analysis to address RQ\textsubscript{pragmatics}. In the end, Section~\ref{sec:discussion} concludes the paper, highlights the limitations of our study, and suggests future directions.

\section{Related Work}
\label{sec:relatedwork}
In this section, we discuss existing studies related to hate and fear speech and show how our work is different from theirs. 

\subsection{Hate Speech}
There is no all-encompassing definition of hate speech \cite{defininghate2014}. But typically, hate speech is considered abusive speech or a direct serious attack on an individual or a group based on the attributes such as race, ethnicity, religion, and sexual orientation \citep{defininghate2016, Hatekeywords2018}. Due to direct and abusive attacks, hate speech often contains toxic words such as \fsl{n*gger} and \fsl{a**hole} that are used against the target. Some provoking sentences (especially involving religious culture)
may directly attack a religious community and can also fall into hate speech. However, many provoking sentences lie outside the scope of hate speech. This is because provoking sentences don't necessarily have toxic words. For example, \emph{``Question- What is non-violence... ?? Answer: Bakra Eid''} uses sarcastic language to provoke readers against the Muslim tradition of killing goats (`Bakra' in Hindi) during the Eid celebration. As a result, this sentence falls under provocation involving religious culture but not hate speech.

Dangerous speech, a subcategory of hate speech, refers to the text that can invoke violence against a group \cite{dangerousspeech2012}. Such violence-invoking cases also overlap with the provocation involving action sentences to some extent. For example, \emph{``Someday Hindus should be ready to fight against Muslims.''} falls into both dangerous speech and provocation involving action. However, provocation involving action is not limited to violence, but also includes cases of supporting anti-sMuslim groups and going against Muslim businesses, organizations, and monuments. For example, \emph{``You buy only from Hindus like all your festivals, etc.''}, asks readers to buy goods only from Hindus and not other religious groups. In India, since Muslims are the next largest religious group after Hindus \cite{Fearspeech2021}, such sentences indirectly urge readers to go against Muslim businesses but don't fall under the umbrella of dangerous speech.

Many studies focus on the automatic identification of hate speech from text \citep{Fearspeech2021, Multilingualhatemodel2021, Multilingualabusivemodel2022, Hatelexicon2018,HS2017,HS2018,HS2016,HS2017-twitter,Hateme2017}. To show that hate speech is different from provocative speech, we test the state-of-the-art hate speech models on \corpus and compare their performance with our transformer-based model (shown in Section~\ref{sec:training}).

\subsection{Fear Speech and Islamophobia}
The dictionary meaning of Islamophobia is fear of, dislike and hate toward, and discrimination against Islam and Muslims \cite{Islamophobia}. However, 
most of the studies on Islamophobia only consider the hate aspect \citep{Detectislamophobia2018,Detectislamophobia2020}. Only a few studies focus on the fear aspect \cite{Fearspeech2021}. \citet{Fearspeech2021} conduct a large-scale study on fearful messages posted on WhatsApp. They curate 27k posts from Indian public WhatsApp groups and find $\sim$\np{8000} of them to be fearful through manual annotation and similarity hashing \cite{Similarity-hashing1999}. Further, they train models to identify such fearful messages. We leverage \citepos{Fearspeech2021} dataset to find provoking sentences that either induce anger or call for some action. We focus on a different emotion (anger instead of fear) evoked against Muslims. We acknowledge that sometimes the same sentence may induce anger and fear in different individuals. In Example~\ref{box:provocation-types}, provoking sentences involving religious culture and oppression may also induce fear among some individuals. However, many provoking sentences such as ``You call Jinnah great \ldots Be ashamed of something'' predominantly instill anger and don't overlap with fear speech. In addition, cases of provocation involving action are not covered under fear speech.

To show that provocation is different than fear speech, we test \citepos{Fearspeech2021} fear speech model 
on \corpus and compare its performance with our transformer-based model (shown in Section~\ref{sec:training}). 

To the best of our knowledge, we are the first ones to study the pragmatics of provocative speech against Muslims, curate a labeled dataset of provoking sentences, and build computational method for the identification.

\section{Answering RQ\textsubscript{identify}}
\label{sec:rq}

\begin{figure}[!htb]
    \centering\includegraphics[width=\columnwidth]{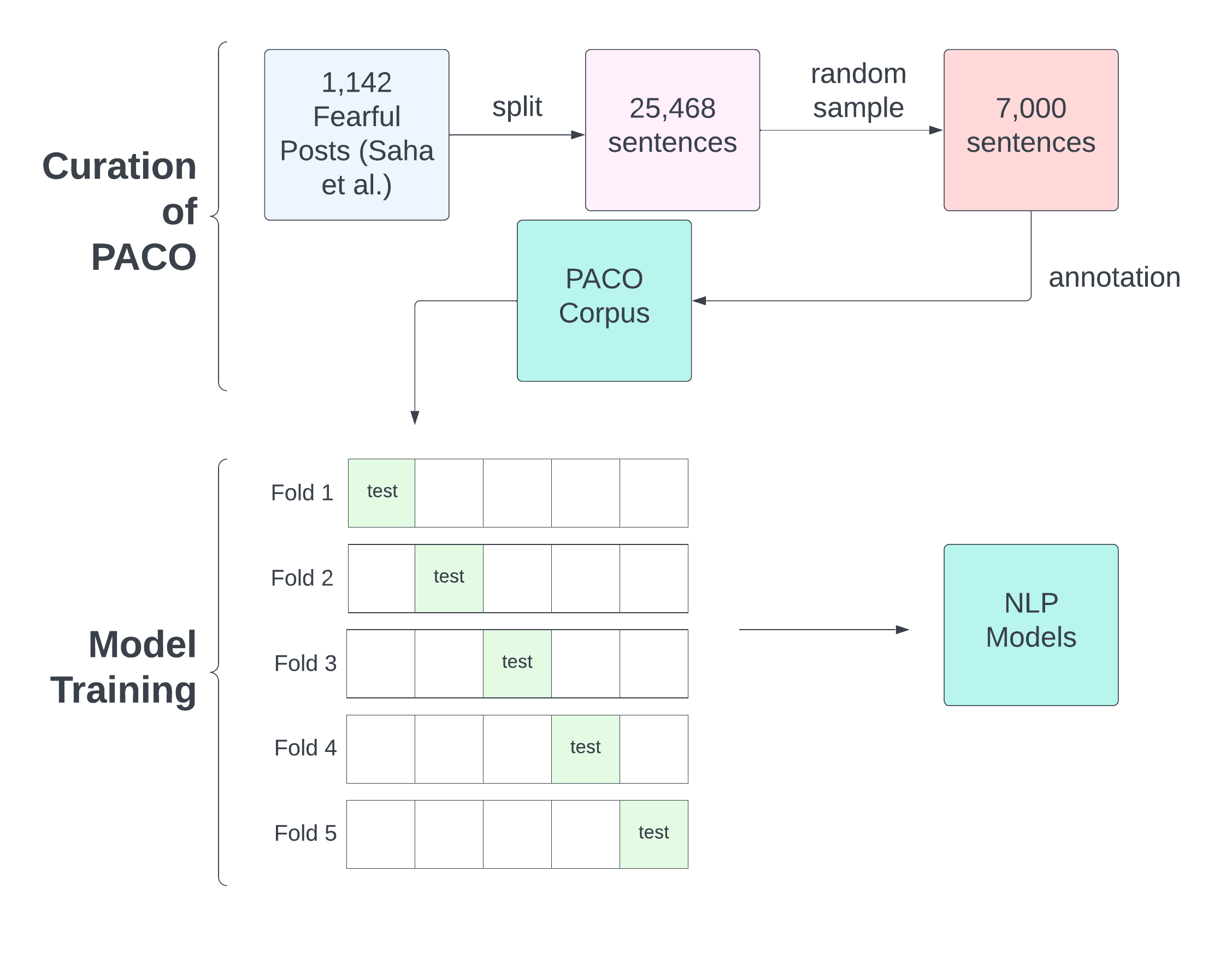}
    \caption{Overview of our method. First, we curated a dataset of provoking sentences called \corpus. Second, we leveraged \corpus to train multiple models and choose the best one for RQ\textsubscript{identify}.}
    \label{fig:overview}
\end{figure}  

Figure~\ref{fig:overview} shows the overview of our method. Our method includes two phases. First, we leveraged fearful posts collected by \citet{Fearspeech2021} and curated \corpus, a dataset of provoking sentences (Section~\ref{sec:dataset}). We also conducted an exploratory data analysis of \corpus (Section~\ref{sec:analysis}). Second, we trained and evaluated multiple embeddings-based and transformer-based models over five-fold cross-validation of \corpus (Section~\ref{sec:training}). Further, we chose the best-performing model for the identification of provoking sentences and compared its performance with the hate speech and fear speech approaches.

\subsection{Curation of \corpus}
\label{sec:dataset}

\citet{Fearspeech2021} identify posts that can instill fear of Muslims, scraped from Indian public WhatsApp groups discussing politics. Out of 27k curated posts, they find $\sim$\np{8000} fearful posts. However, they only shared \np{4782} posts publicly, out of which \np{1142} posts were fearful. As indicated in Example~\ref{box:fear-provocation}, fearful posts also contain provoking sentences. Hence, we split \np{1142} fearful posts into \np{25468} sentences (using the sentence tokenizer of Natural Language Toolkit (NLTK) library \cite{Nltk}) and randomly sampled \np{7000} sentences for annotation purposes.

The first two authors of this paper were the annotators. They were given a sentence and asked to label one of the following types: (i) provocation involving culture, (ii) provocation involving oppression, (iii) provocation involving action, and (iv) none. Along with the sentence to annotate, they were provided its preceding and succeeding sentences to get enough context while labeling.  

The annotation was conducted in three phases. In the first phase, the annotators were provided with initial labeling instructions, which they used to label a total of 400 sentences, in four rounds of 100 sentences each. In this phase, Cohen's kappa score came out to be 0.48 (moderate agreement) \cite{Cohen1960}. After each round, the annotators discussed their disagreements, which helped in finalizing the labeling instructions. In the second phase, the two annotators labeled 600 sentences using the final instructions, leading to Cohen's kappa score of 0.73 (substantial agreement). In the third phase, the remaining \np{6000} sampled sentences were split among the two annotators, such that only one annotator labeled a sentence.

The final labeling instructions, including definitions and examples of each provocation category, are shown below:
\begin{itemize}
    \item \textbf{Action:} Sentences that urge readers to act against a religious group were labeled as provocation involving action. For example, the following sentence asks readers to support Hindus in the fight against Muslims. This support-seeking is open to interpretation and can mean anything from boycotts to violence against Muslims. Thus, we considered such cases in this category. \bigskip
    
    \emph{``Always support Hindus in the fight of Hindu Muslim, right wrong no matter, all of them later!''}\bigskip

    We also considered subtle cases that indicate some action. For example, an indirect call to shut down a madrasa (Islamic school) is depicted in the following:\bigskip

    \emph{``If the madrasa is not closed, after 15 years, more than half of the Muslims of the country will be supportive of the ISIS ideology''}\bigskip

     \item \textbf{Religious oppression:} Sentences that express oppressing incidents (real or fake) as facts, to provoke readers against a particular religious group were labeled as provocation involving religious oppression. A relevant sentence structure is ``X did violence to Y'', where X is Muslim(s) in our use case and Y is an individual or some other group of people. Such an example is presented below.\bigskip 

     \emph{``<A person> - set fire - heavy damage at Sartala; 3 out of four rooms were destroyed''}\bigskip
     
    \item \textbf{Religious culture:} Sentences that can make readers annoyed at a religion's sacred books, leaders, or people of that religion, were labeled relevant for this category. Following are a few examples:\bigskip 

    ``In Kashmir, every person who speaks `Murdabad' is a Muslim'' (targeting Muslims)\bigskip

    ``\ldots Muslim children are well taught Jihadi Quran in madrasa''
    (targeting Quran, the main Islamic scripture)\bigskip

    The first example above targets all Muslims that they say 'Murdabad' in Kashmir (meaning India's dismissal). Moreover, the second example mentions the Quran to be Jihadi, meaning it spreads terrorism. Indirectly, the latter example typecasts all Muslim children to become terrorists. Such sentences invoke anger against a religious group and hence are sconsidered provoking.  
\end{itemize}
The sentences not indicating any of the above types were labeled `none'.
Combining the sentences annotated in the three phases, we curated a dataset of \np{7000} sentences. We call this dataset \corpus.

\textbf{Ethics note:} We annotated the sentences present in fearful posts. Such fearful posts were shared by \citet{Fearspeech2021} and did not include any private information of WhatsApp users who wrote them. Moreover, since these posts were part of public WhatsApp groups, neither \citet{Fearspeech2021} nor us were required to take the consent of WhatsApp users before using the text. We acknowledge that the nature of the text can be disturbing, especially for Muslims. That's why we did not hire crowd workers to annotate sentences. Instead, the authors of this paper who were aware of the nature of the text completed the annotation.

\subsection{Exploratory Data Analysis}
\label{sec:analysis}

\begin{figure}[!htb]
    \centering\includegraphics[width=0.7\columnwidth]{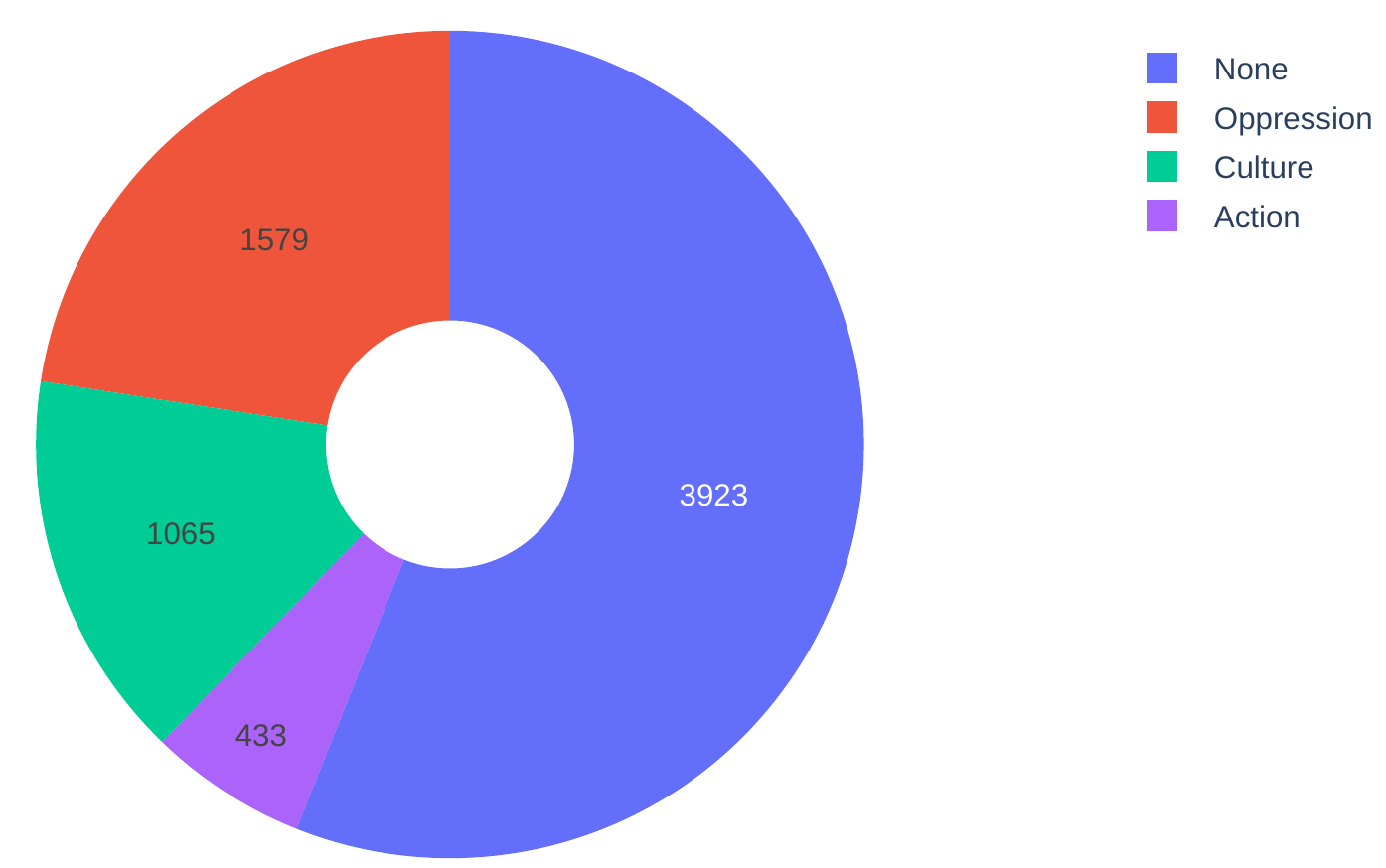}
    \caption{Donut chart showing the number of sentences in each of the four classes.}
    \label{fig:distribution}
\end{figure}

Figure~\ref{fig:distribution} shows the distribution of the annotated sentences across the four classes: (i) action, (ii) religious oppression, (iii) religious culture, and (iv) none. Out of \np{7000} sentences, \np{433} (6.19\%) were provoking sentences involving an action, \np{1065} (15.21\%) used religious culture, and \np{1579} (22.55\%) used religious oppression. Moreover, there were \np{3923} (56.05\%) sentences that did not belong to any of these provocation types (none category). 

We also visualized word clouds for each of the provocation categories to know the most frequent words in them. In all three categories, words such as \fsl{Hindus} and \fsl{Muslims} were among the most frequent words. To see other frequent words, we removed the words, \fsl{Hindus} and \fsl{Muslims}, from the text and visualized the word clouds.

\begin{figure}[!htb]
    \centering\includegraphics[width=0.9\columnwidth]{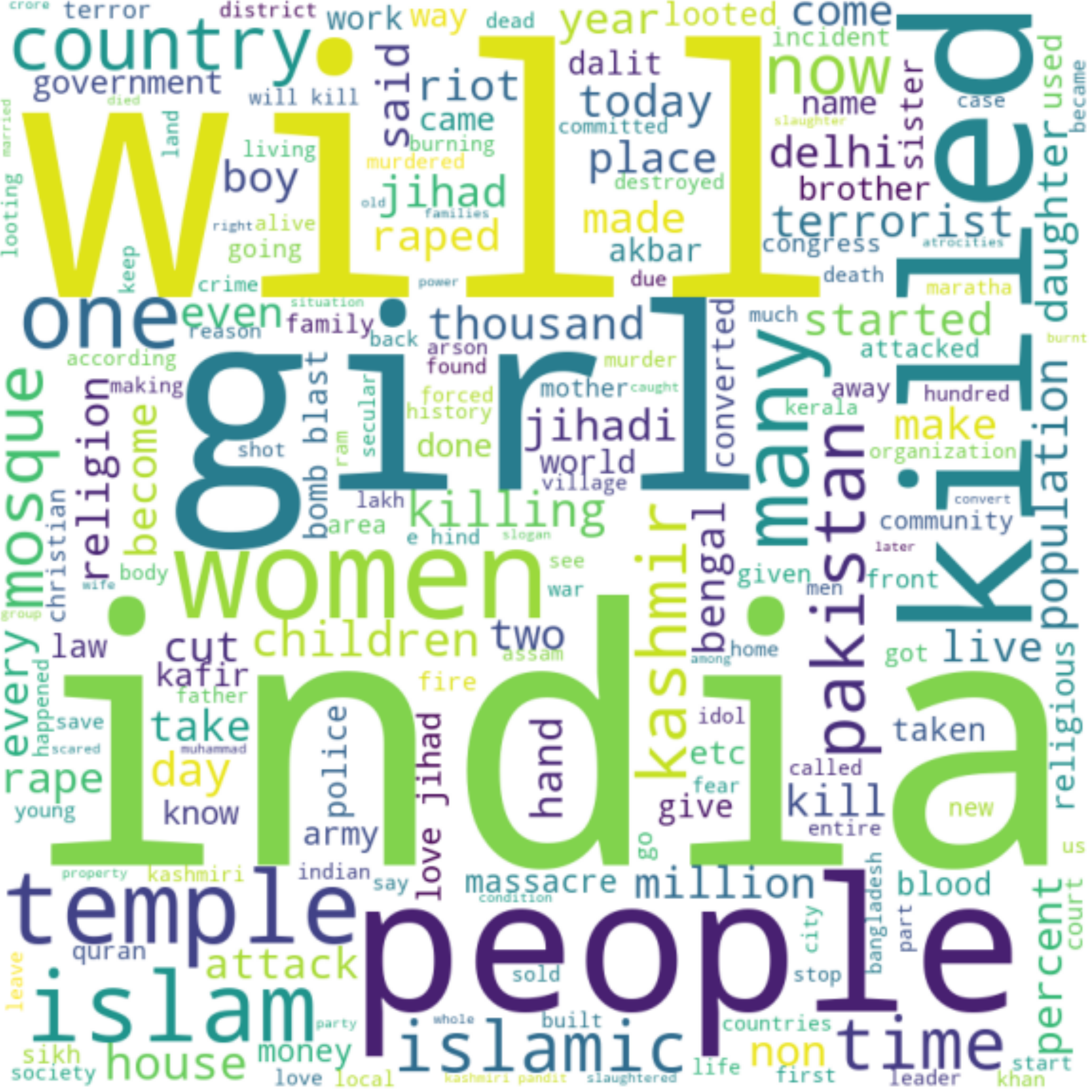}
    \caption{Word cloud for sentences of the religious oppression category.}
    \label{fig:oppression}
\end{figure}

Figure~\ref{fig:oppression} shows the word cloud for the religious oppression category. As this category describes past oppressive incidents, the word cloud shows words related to the victims of such oppression such as \fsl{girl}, \fsl{women}, \fsl{daughter}, \fsl{kafir}, and \fsl{children}.  Moreover, it is showing words that describe the act of oppression such as \fsl{killed}, \fsl{raped}, \fsl{cut}, \fsl{riot}, \fsl{terrorist}, \fsl{jihadi}, and \fsl{bomb}.

\begin{figure}[!htb]
    \centering\includegraphics[width=0.9\columnwidth]{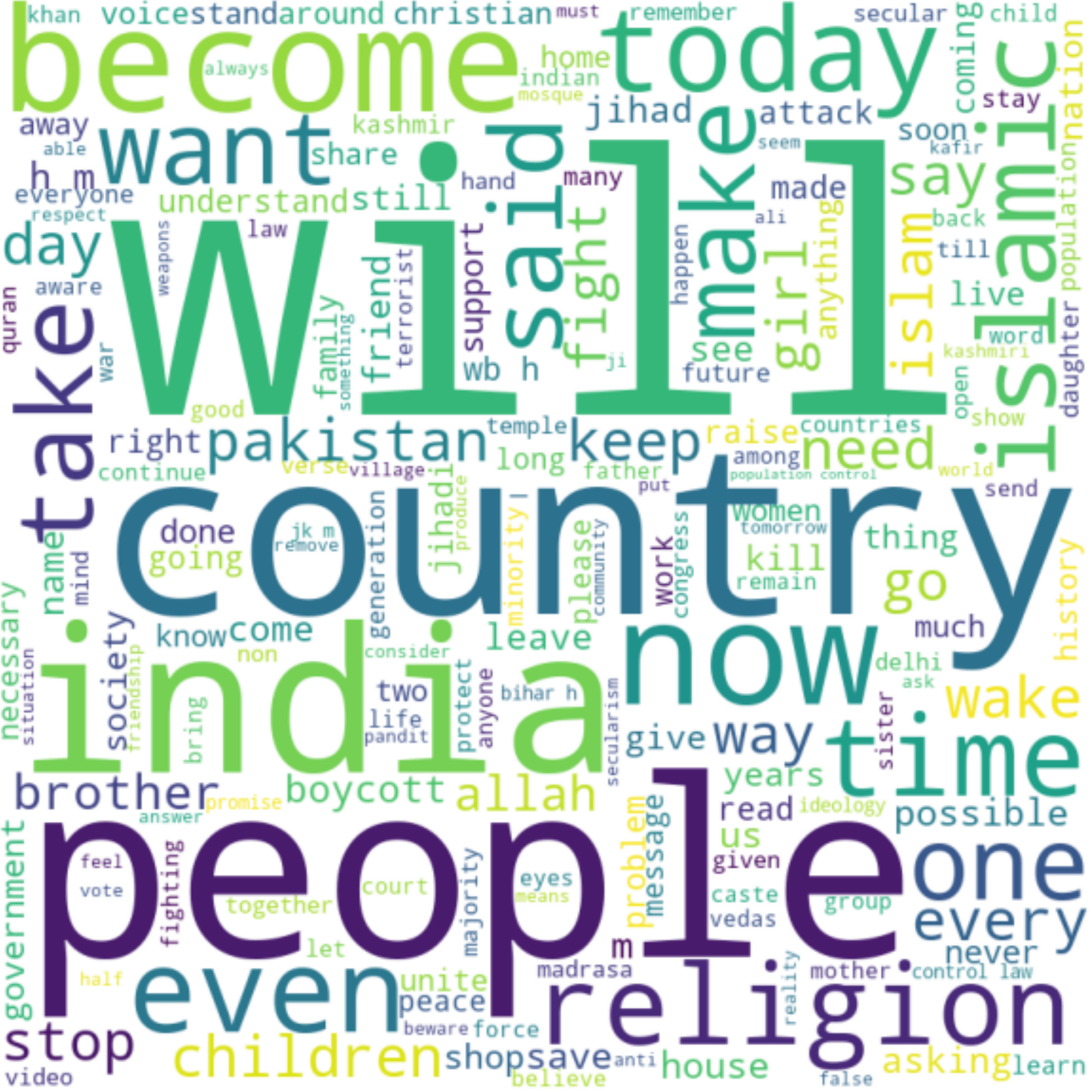}
    \caption{Word cloud for sentences of the action category.}
    \label{fig:action}
\end{figure}

Figure~\ref{fig:action} shows the word cloud for the action category. Words such as \fsl{country} and \fsl{India} are prominent. This is because such sentences ask readers to eradicate Muslims (or their possessions) because of their negative impact on the `country' or `India'. In addition, we observed many action-describing words such as \fsl{make}, \fsl{raise}, \fsl{wake}, \fsl{voice}, \fsl{stand}, and \fsl{fight}.

\begin{figure}[!htb]
    \centering\includegraphics[width=0.9\columnwidth]{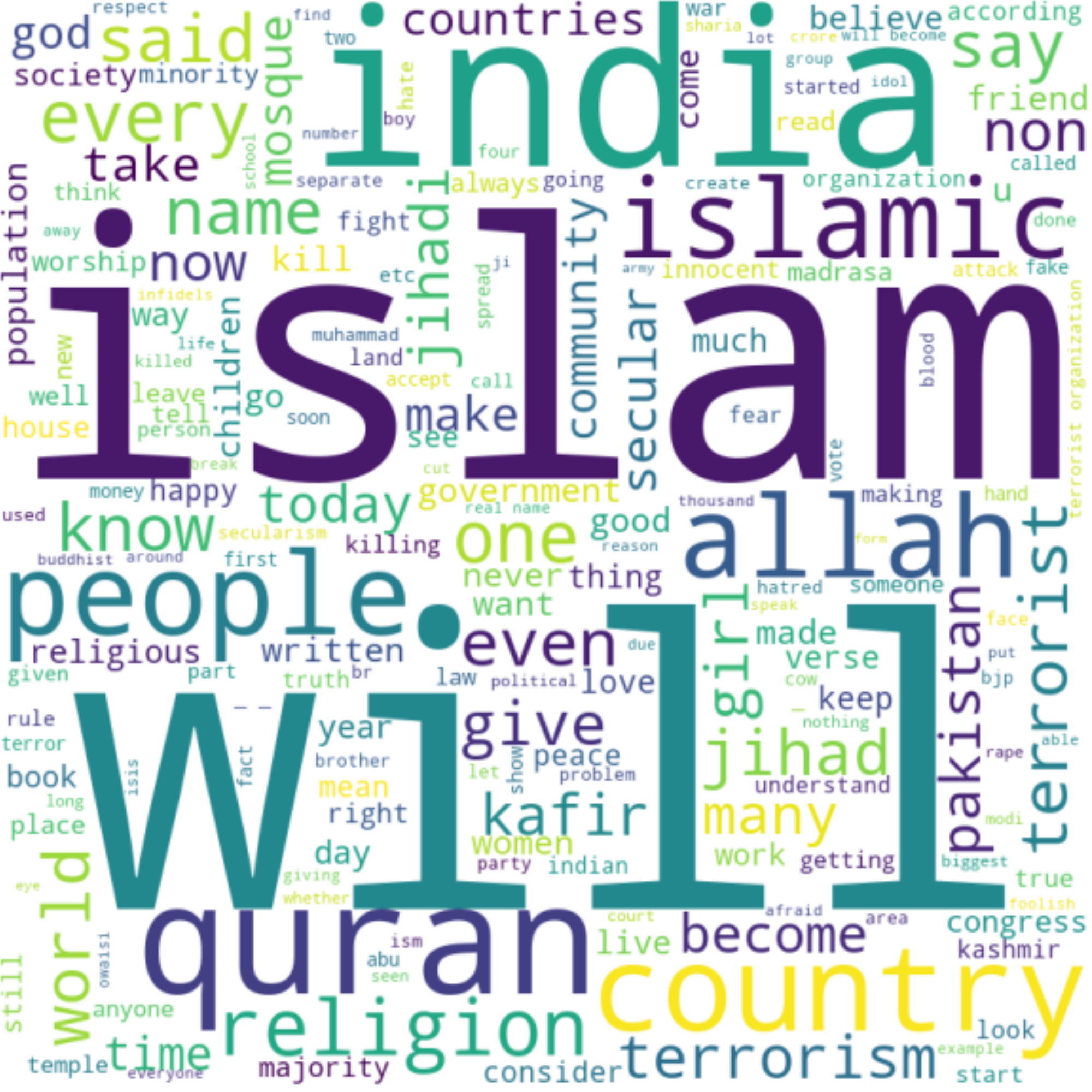}
    \caption{Word cloud for sentences of the religious culture category.}
    \label{fig:culture}
\end{figure}

Figure~\ref{fig:culture} shows the word cloud for the religious culture category. It has many words describing the target religious group such as \fsl{islam}, \fsl{islamic}, \fsl{quran}, and \fsl{allah}. In addition, it shows some words that are also present in the oppression word cloud, such as \fsl{jihad}, \fsl{jihadi}, and \fsl{terrorism}. These words have a different context for the culture category, basically used to stereotype the people of a religious group.

Moreover, there are words such as \fsl{people}, \fsl{will}, and \fsl{said} that are common in all word clouds. 

\subsection{Model Training}
\label{sec:training}

We considered our problem a multiclass classification task. For multiclass classification, we 
explored multiple training approaches on \corpus. Each sentence was input to a model and the output was one of the four classes: (i) religious culture, (ii) religious oppression, (iii) action, and (iv) none. We discuss multiple approaches and compare their performances below.

\begin{description}
    \item[TF-IDF] weighs each word in the corpus according to its Term Frequency (TF) and Inverse Document Frequency (IDF) \cite{Tfidf2021}. Based on the number of unique tokens in \corpus, TF-IDF yielded a \np{11441}-dimensional embedding for each sentence of this dataset. We provided these embeddings as input to multiple classifiers such as Logistic Regression (LR) \cite{LR2002}, Random Forest (RF) \cite{RF2014}, and Support Vector Machine (SVM) \cite{SVM2020}, and compared their performance.
    
    \item[Word2Vec] converted each word in \corpus into a \np{300}-dimensional embedding \cite{Word2Vec2013}. In our case, we obtained such word embeddings using the Word2Vec model pretrained on the Google News dataset. To form a sentence embedding, we averaged Word2Vec embeddings for all words present in that sentence. Further, sentence embeddings were provided as inputs to LR, RF, and SVM.
    
    \item[GloVe] yields an embedding for each word in the corpus \cite{Glove2014}. We used Stanford's GloVe model, which is trained on the Wikipedia dataset and returns a \np{100}-dimensional word embedding.
    We averaged these word embeddings in the same way as we did in Word2Vec. Finally, we trained the above three classifiers on GloVe sentence embeddings.
    
    \item[Universal Sentence Encoder (USE)] leverages Deep Averaging Network (DAN) to extract \np{512}-dimensional embeddings for each sentence \cite{UniversalSentenceEncoder2018}. We leveraged these embeddings as features of the above classifiers.

    \item[Transformer-based models:] We leveraged modern transformer-based approaches such as BERT \cite{BERT2019}, RoBERTa \cite{Roberta2019}, and XLNet \cite{Xlnet2019}. We fine-tuned all these models on our dataset by adding a layer (with softmax activation) in the forward direction, containing four output units (one for each class). Further, these models were trained using a batch size of \np{32}, the maximum sequence length of \np{256}, for five epochs to minimize the cross-entropy loss.
\end{description}

All the above approaches were evaluated on \corpus for five-fold cross-validation. Moreover, \corpus was divided into five folds in a stratified manner, leading to the same class distribution in each fold. For evaluation, prior works on hate and fear speech \citep{Fearspeech2021, Twitterhate2019, Onlinehate2020} leveraged the AUC-ROC metric because it measures the goodness of fit, especially appropriate for imbalanced datasets. \corpus also suffers from imbalanced class distribution (Figure~\ref{fig:distribution}). Hence, we also leveraged AUC-ROC score for evaluation.

Table~\ref{tab:results} shows AUC-ROC score (obtained by one versus one method \cite{Onevsone}) achieved by each of the above approaches in five folds. Among embeddings-based methods, TF-IDF with LR achieved 0.818 as the average AUC-ROC score, followed by Word2Vec with SVM (0.815), USE with SVM (0.812), and GloVe with SVM (0.800). However, all transformer-based approaches such as BERT (0.832 average AUC-ROC), RoBERTa (0.851 average AUC-ROC), and XLNet (0.845 average AUC-ROC) outperformed embedding-based approaches. Overall, RoBERTa achieved the highest average AUC-ROC score and was chosen as the best model.

On \corpus, we checked the performance of existing approaches \citep{Fearspeech2021,Hatelexicon2018, Multilingualhatemodel2021, Multilingualabusivemodel2022} for fear and hate speech. To do so, we followed three steps. First, in \corpus, we considered sentences of three provoking categories (action, religious culture, and religious oppression) as relevant (positive class) and others as irrelevant (none category as the negative class). Second, we leveraged the hate lexicon-based method \cite{Hatelexicon2018}, state-of-the-art hate speech models \cite{Multilingualhatemodel2021, Multilingualabusivemodel2022}, and fear speech model \cite{Fearspeech2021} to identify relevant sentences from \corpus. We implemented \citepos{Fearspeech2021} fear speech model using the Simple Transformers library \cite{Simpletransformers} with the same hyperparameters that they used. Moreover, hate speech models \citep{Multilingualhatemodel2021, Multilingualabusivemodel2022} were available on the Hugging Face platform \citep{De-limit, Abusivespeech} and \citepos{Hatelexicon2018} hate lexicons were available on their Github repository \cite{Hatelexicons}. Third, we computed the AUC-ROC score for each of these approaches.

Table~\ref{tab:baselines} shows the AUC-ROC scores of hate and fear speech models. Among them, the highest AUC-ROC score was achieved by \citepos{Multilingualabusivemodel2022} hate speech model (0.726), followed by \citepos{Multilingualhatemodel2021} model (0.668), \citepos{Fearspeech2021} fear identifying approach (0.643), and \citepos{Hatelexicon2018} hate speech lexicons (0.602). In identifying provoking sentences, all these approaches perform worse than our RoBERTa model (average AUC-ROC score of 0.851 on five-fold cross-validation). Moreover, even while using these hate or fear speech models, there is no means of knowing the category of provocation, as opposed to our RoBERTa model.

\begin{table}[!htb]
    \centering
    \caption{AUC-ROC score for each of the five folds. Bold indicates the highest average AUC-ROC score among all approaches.}
    \label{tab:results}    
\resizebox{\columnwidth}{!}{\begin{tabular}{l c c c c c c}
    \toprule
        Approach & Fold 1& Fold 2 & Fold 3& Fold 4& Fold 5 & Average\\
        \midrule
        TF-IDF + LR & 0.810& 0.825& 0.832& 0.805& 0.820 & 0.818\\
        TF-IDF + RF & 0.790& 0.793& 0.812& 0.786& 0.803 & 0.780\\
        TF-IDF + SVM & 0.810 & 0.818& 0.832& 0.801& 0.812 & 0.814\\
        \midrule
        Word2Vec + LR & 0.810 & 0.800 & 0.788& 0.792& 0.784 & 0.794\\
        Word2Vec + RF & 0.756 & 0.752 & 0.747& 0.745& 0.758 & 0.751\\
        Word2Vec + SVM & 0.823 & 0.820& 0.805& 0.816& 0.814 & 0.815\\
        \midrule
        GloVe + LR & 0.784 & 0.785& 0.775& 0.755& 0.766 & 0.773\\
        GloVe + RF & 0.760& 0.758& 0.752& 0.760& 0.758 & 0.757\\
        GloVe + SVM & 0.809& 0.812& 0.793& 0.794& 0.795 & 0.800\\
        \midrule
        USE + LR & 0.812 & 0.825& 0.801& 0.786& 0.807& 0.806\\
        USE + RF & 0.768 & 0.773& 0.760& 0.754& 0.757& 0.762\\
        USE + SVM & 0.821 & 0.829& 0.811& 0.798& 0.803& 0.812\\
        \midrule
        BERT & 0.838 & 0.826& 0.839& 0.819& 0.838& 0.832\\
        RoBERTa & 0.851 & 0.849& 0.854& 0.845& 0.856 & \textbf{0.851}\\
        XLNet & 0.843 & 0.850 & 0.856& 0.833& 0.847 & 0.845\\
    \bottomrule
    \end{tabular}
    }
\end{table}

\begin{table}[!htb]
    \centering
    \caption{Performance of hate and fear speech approaches on \corpus.}
    \label{tab:baselines}    
    \begin{tabular}{c c}
    \toprule
        Approach & AUC-ROC\\
     \midrule
        Hate speech lexicons (\citet{Hatelexicon2018})& 0.602 \\
        Fear speech model (\citet{Fearspeech2021})&  0.643\\
        Hate speech model 1 (\citet{Multilingualhatemodel2021}) & 0.668 \\
        Hate speech model 2 (\citet{Multilingualabusivemodel2022}) & 0.726 \\
    \bottomrule
    \end{tabular}
\end{table}

\section{Answering RQ\textsubscript{pragmatics}: Qualitative Analysis}
\label{sec:rq2}
We randomly selected \np{300} provoking sentences (\np{100} from each category) from \corpus and read them to identify dialogue acts specific to religious provocation. For \np{200} of them (sentences of religious culture and oppression), we even identified impoliteness super-strategies used in the text.

\subsection{Dialog Acts}
\citet{Speechacts1962} 
defined dialogue acts as the function of the speaker's utterance. Following \citepos{Speechacts1962} work, \citet{Speechacts1969} identified five types of dialog acts that are generally used in natural language. However, recent social-media research emphasizes on identifying dialog acts that are specific to the scenario \citep{Metooma2020,Twitterdialog2015}. For the provocation (against Muslims) scenario, we identified the following dialog acts.

\begin{description}
    \item[Accusation:] Sentences that blame Muslims or their leaders for a specific negative event in past. Some examples are listed below.\bigskip

    \emph{``Entered, broke the Shivling into pieces, and acquired as much property as he could in the solution''}\bigskip

    \emph{``Muslim doctor silently sterilization of 4000 Buddhist women \ldots''}\bigskip

    \item[Defaming Muslims:] Sentences that target Muslims for their present behavior or generalize them with a negative trait. For example,\bigskip 

    \emph{``So Pakistan and Kashmiri Muslims take a blind eye to the atrocities on Uyghurs in Xinjiang \ldots''}\bigskip

    The above sentence targets Muslims' present behavior for not showing support to Uyghurs.\bigskip 
    
    \emph{``Now you can think that whom will these Muslims serve as IAS \ldots Islam or country?''}\bigskip
    
    The above rhetorical question generalizes that all Muslims support Islam more than India. Defaming generalizes Muslims as opposed to accusation that is made over a specific event.\bigskip
    
    \item[Criticizing Islam:] Sentences that either find loopholes in Islam or completely dismiss it. For example, the following sentence criticizes Islam by pointing out that a non-Muslim's (also called kafir) statement does not hold in Islamic court (Sharia).\bigskip

    \emph{``The testimony of Kafir in Sharia court, ie Qazi court, is not valid''}\bigskip

    Moreover, sentences such as following completely dismiss Islam by calling it `anti-human religion'.\bigskip
        
    \emph{``\ldots Which superstitious religions are in India * * Yes Islam is the anti-human religion of superstition * * 1) If Allah , God is equally powerful * * There is God and he hates Kafiro * * And Kafiro does not have the right to live''}\bigskip

    \item[Comparison:] Sentences that compare Muslims, their emperors, or traditions with those of other religious groups.\bigskip

    \emph{``You keep teaching your children that Akbar was great but never think of yourself then what to say to that Rana Pratap, who wandered in the forests all his life, fighting with Akbar even after eating grass bread''}\bigskip

    In the above example, Akbar who was a Muslim emperor is compared with Rana Pratap, a Hindu emperor.\bigskip

    \emph{``Today, innocent Hindus and saints are in jail and ungodly people are out today''}\bigskip

    \item[Commanding:] Sentences that directly call for an action. Below are some examples.\bigskip

    \emph{``Be violent for religion''}\bigskip

    \emph{``To save the respect of all Hindu countrymen and protect your security, keep weapons in your house''}\bigskip

    \item[Motivating:] Sentences that indirectly urge to take an action.\bigskip

    \emph{``If the madrasa is not closed, after 15 years, more than half of the Muslims of the country will be supportive of the ISIS ideology''}\bigskip

    The above sentence claims that Muslims will join ISIS (a terrorist organization) if madrasas (Islamic schools) are not closed. Indirectly, this sentence motivates to shut these Islamic schools.
\end{description}

For religious culture and oppression categories, three dialog acts were widely used: accusation, defaming Muslims, and criticizing Islam. In the action category, commanding and motivating were the most frequent. For comparison, we found only a few cases that were distributed among religious culture and oppression categories.

\subsection{Impoliteness Super-Strategies}
\citet{Impolitenessdefn2011} defined impoliteness as a negative attitude toward specific behaviors in specific situations. Such negative attitude can lead to emotional consequences within the target \cite{Impolitenessdefn2011}. Since provoking sentences involving religious culture and oppression attack or accuse Muslims or Islam, they are impolite to Muslim readers. However, provoking sentences in the action category don't verbally attack Muslims or Islam. Hence, for impoliteness analysis, we only considered randomly selected sentences from the religious culture and oppression category.

\citepos{Impolitenessmodel2003} model describes the following five super-strategies for impoliteness.

\begin{description}
    \item[Bald on record] is used when there is a direct attack to the face of the target. 
    \item[Positive impoliteness] is used to destroy the positive face of the target. It includes the cases of ignoring or excluding them, being unsympathetic and unconcerned toward them, making them uncomfortable, and using obscure and taboo language.
    \item[Negative impoliteness] destroys the negative face of the target. It includes cases of associating the target with a negative aspect, condescend, frighten, or ridicule them, and invading their space. 
    \item[Sarcasm] is used to indirectly say opposite to the literal meaning of the text.
    \item[Withhold politeness] refers to the scenario where politeness is expected of the speaker but they become silent or fail to act. For example, forgetting to say thanks after the other person helps you.  
\end{description}
We analyzed the randomly chosen sentences (from religious culture and oppression categories) to find which of these super-strategies are commonly used.
Since we deal with individual sentences and not the dynamics between posts, withhold politeness is ruled out. Moreover, we did not find obscure, taboo, or ignoring language showing positive impoliteness. 

We found only one sentence using sarcasm, which targets Muslims for claiming that they are a minority, ``India has the largest Muslim population in the world after Indonesia ???? Oddly enough, it is still a minority ??????''. According to our analysis, sarcasm is not widely used across provoking sentences.

Predominantly, two super-strategies: bald on record and negative impoliteness were found in the provocative speech. For example, \emph{``Muslims are not friends of anyone''} directly attacks Muslims (bald on record). For negative impoliteness, we found cases that associate Muslims or Islam with a negative aspect. But such cases are indirect unlike bald on record. For example, \emph{``Alauddin Khilji summoned Rana Ratan Singh of Chittor on the pretext of friendship and then killed''}, associates Alauddin Khilji (a Muslim emperor) with a negative aspect for killing Rana Ratan Singh (a Hindu emperor). This sentence with the whole context indirectly infers that Muslims always oppress Hindus.

\section{Discussion}
\label{sec:discussion}
We now discuss our conclusion, limitations of our study, and propose future directions of research.

\subsection{Conclusion}
Prior studies \citep{Hatekeywords2018, Violenceprovocation2013, Fearspeech2021} focus on identifying hate and fear speech. To the best of our knowledge, there is no existing study on identifying provocative speech in the religious context. We aimed to identify provoking sentences against Indian Muslims.
We labeled a dataset of \np{7000} provoking sentences for the three categories: action, religious culture, and religious oppression. We call this dataset \corpus. To solve the identification problem, we leverage \corpus to train and evaluate multiple NLP models (embedding-based and transformer-based) over five-fold cross-validation. Our best-performing model, RoBERTa achieves an average of 0.851 AUC score. The automatic identification of provoking sentences can prevent the spread of provocative speech and in turn prevent possible violence against the target religious group.

Moreover, we studied the provocative text against Muslims through the pragmatic lens and identified the dialog acts and the impoliteness super-strategies used.

\subsection{Limitations and Future Work}
First, our work is specific to identifying sentences that provoke readers against Indian Muslims. In the future, we can expand our study to identify provoking sentences against other religious groups such as Hindus, Sikhs, and so on. Second, we leveraged sentences from only one social media platform, WhatsApp. We can expand \corpus to include sentences from multiple platforms (such as Reddit and Twitter) and train cross-platform-based models for identification. Further, NLP models can be built with a broad vision of identifying all three: provocative speech, fear speech, and hate speech. Such models will serve 
as a one-stop solution to 
eliminate all the disturbing and targeted text from social media platforms.

\bibliographystyle{IEEEtranSN}
\bibliography{Vaibhav}

\end{document}